\documentclass[runningheads]{llncs}
\usepackage[T1]{fontenc}
\usepackage{graphicx}
\usepackage{multirow}
\usepackage{hyperref}
\usepackage{booktabs}
\usepackage{wrapfig}
\DeclareUnicodeCharacter{2212}{-}
\makeatletter
\def\@fnsymbol#1{\ensuremath{\ifcase#1\or \dagger\or \ddagger\or
   \mathsection\or \mathparagraph\or \|\or **\or \dagger\dagger
   \or \ddagger\ddagger \else\@ctrerr\fi}}
\makeatother

\begin{document}
\title{On-the-Fly Guidance Training for Medical Image Registration}

\author{
Yuelin Xin\inst{1,2}* \thanks{Corresponding author}
\and
Yicheng Chen\inst{3,5}*
\and
Shengxiang Ji\inst{4,6}*
\and
\\Kun Han\inst{2}*
\and
Xiaohui Xie\inst{2}
}

\authorrunning{Y. Xin et al.}

\institute{
\mbox{
\inst{1}University of Leeds
\quad
\inst{2}University of California, Irvine
\quad
}
\mbox{
\inst{3}Tongji University
\quad
\inst{4}Huazhong University of Science and Technology
}
\mbox{
\inst{5}Fudan Univeristy
\quad
\inst{6}University of California, San Diego
}
}

\maketitle              

\begin{abstract}
This study introduces a novel \textbf{On-the-Fly Guidance} (OFG) training framework for enhancing existing learning-based image registration models, addressing the limitations of weakly-supervised and unsupervised methods. Weakly-supervised methods struggle due to the scarcity of labeled data, and unsupervised methods directly depend on image similarity metrics for accuracy. Our method proposes a supervised fashion for training registration models, without the need for any labeled data. OFG generates pseudo-ground truth during training by refining deformation predictions with a differentiable optimizer, enabling direct supervised learning. OFG optimizes deformation predictions efficiently, improving the performance of registration models without sacrificing inference speed. Our method is tested across several benchmark datasets and leading models, it significantly enhanced performance, providing a plug-and-play solution for training learning-based registration models. Code available at: \url{https://github.com/cilix-ai/on-the-fly-guidance}

\keywords{image registration \and on-the-fly guidance \and pseudo label.}
\end{abstract}

\section{Introduction}
Medical image registration is pivotal in medical image analysis, aiming to align two medical images by optimizing their visual similarity through a deformation field. There are two factions: traditional optimization-based methods like \cite{AVANTS200826,BAJCSY19891,1175091}, which iteratively refine the deformation field using mathematical constraints, and modern learning-based methods \cite{Chen_2022,chen2021vitvnet}, which predict deformation fields from image pairs. Both approaches are vital, with the latter gaining significant traction for its direct prediction capabilities, marking a swift evolution in the field.

Nevertheless, learning-based methods face a major hurdle: the trade-offs between weakly-supervised learning, which yields better results at the cost of extensive labeling, and unsupervised learning, which foregoes labels but directly relies on less precise image similarities for deformation field derivation. This situation prompts the critical question: \textit{is it possible to create a training method that bypasses the need for manual labels while still benefiting from the precision of direct supervision?}

In this work, we present a novel training framework named on-the-fly guidance (OFG) that merges supervised learning with existing learning-based image registration methods to boost their performance. OFG uniquely generates pseudo-ground truth on the fly through instance-specific optimization, using these results for direct supervision. This hybrid approach combines direct prediction with iterative refinement in a two-stage process: 1) the model predicts a deformation field $\phi_{pre}$, and 2) this field is optimized to produce $\phi_{opt}$, which then acts as a pseudo-label. The model's training is guided by directly comparing the initial and optimized deformation fields using $MSE(\phi_{pre}, \phi_{opt})$, ensuring a direct and efficient learning process.

OFG introduces a crucial incremental supervision method that guides models toward convergence by setting intermediate goals rather than a final objective. It optimizes the current deformation prediction in a step-by-step manner, easing the model's learning process while balancing supervision quality with computational efficiency. This approach allows image registration models to undergo a nuanced, self-improving training process. Compared to baseline unsupervised and other pseudo-supervised methods like self-training, OFG shows consistent and significant improvements, underscoring its effectiveness.

The main contributions of our work are summarized as follows:
\begin{itemize}
    \item Introducing OFG, a training framework that enhances existing image registration models, utilizing supervised learning without relying on labeled data.
    \item Using optimized pseudo-ground truth for incremental learning targets, fostering a self-enhancing cycle between the model and optimizer.
    \item Presenting through extensive benchmarks that our method surpasses baselines and previous state-of-the-art across different datasets and models.
\end{itemize}

\section{Related Work}
\label{sec:related_work}

\textbf{Weakly-Supervised \& Unsupervised Training.} Learning-based methods have recently overtaken traditional optimization-based approaches \cite{10.1007/978-3-540-30135-6_78,patch-based,GLOCKER2008731,THIRION1998243,lddmm} in performance and efficiency \cite{Lea2022IEEE,Aco2018arX,ale2021CoRR,are2022Fro}, with supervised methods \cite{yang2017quicksilver,10.1007/978-3-319-66182-7_27} depending on ground-truth deformation fields often derived from traditional techniques. The rise of unsupervised methods \cite{Balakrishnan_2019,de_Vos_2019,shen2019netw,zhang2021lear,chen2021vitvnet,Chen_2022} optimize metrics like NCC to understand the dataset globally. Popular unsupervised models like VoxelMorph \cite{Balakrishnan_2019}, ViT-V-Net \cite{chen2021vitvnet}, and TransMorph \cite{Chen_2022} employ a U-Net structure with CNN or ViT elements for deformation prediction and can use segmentation labels for enhanced accuracy, though with the high cost of annotations.

\textbf{Self-training.} Closely related to our research is Cyclical Self-training \cite{uns2023MICCAI}, which adopts a teacher-student approach, alternating training stages and employing pseudo labels for guidance. Our approach differs significantly in the following ways: \textbf{1)} the optimizer in \cite{uns2023MICCAI} employs a non-learnable approach to generate deformation from the fusion of two encoded image features, which is not competitive compared with existing learning-based models, \textbf{2)} OFG generates pseudo labels incrementally for each training epoch, as opposed to updating labels between training stages which may introduce challenging shifts for the model, and \textbf{3)} OFG offers an end-to-end, plug-and-play framework applicable across different models and datasets, contrasting with the custom, less generalizable approach of Cyclical Self-training.

\begin{figure*}[t]
    \begin{center}
      \includegraphics[width=\linewidth]{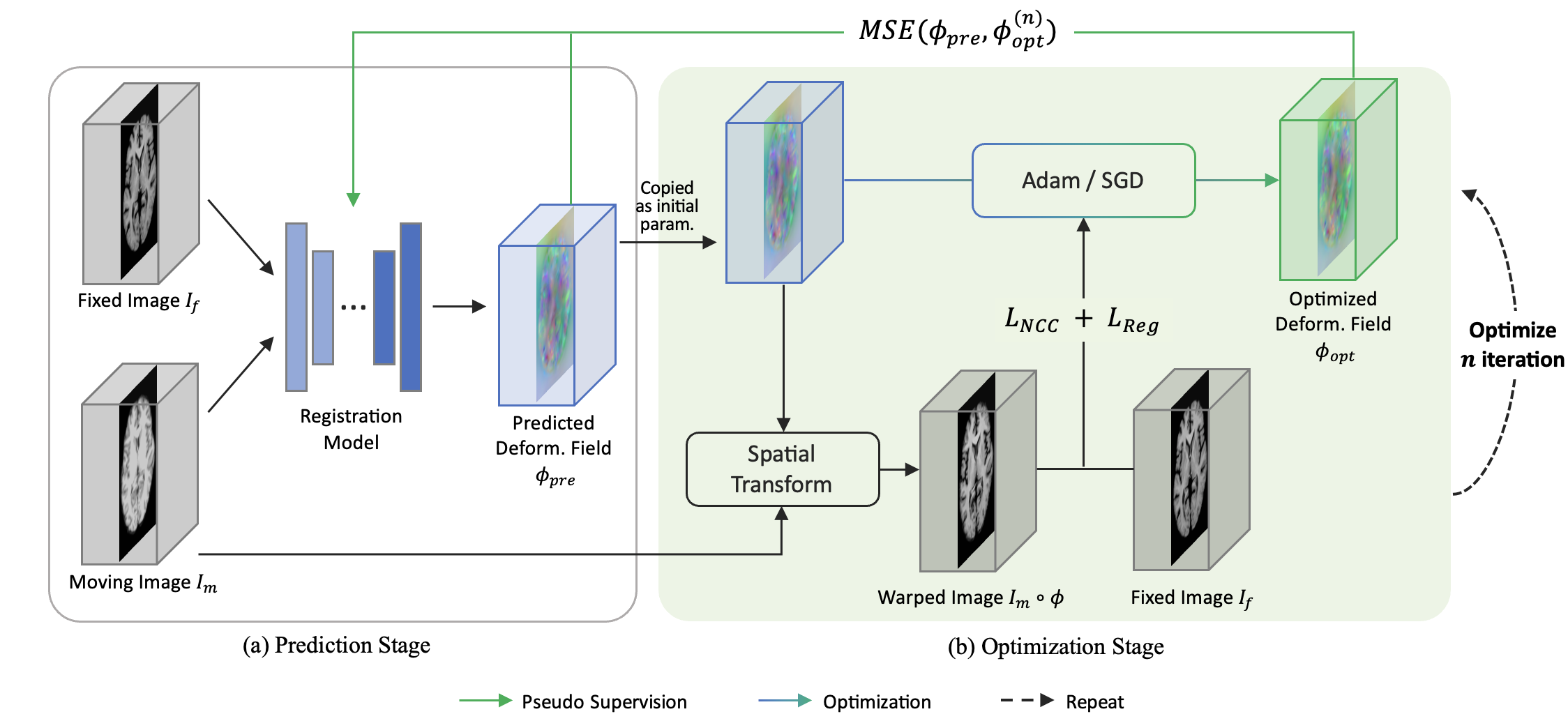}
    \end{center}
    \caption{\small The overall structure of the proposed framework. It has two parts, the prediction stage (a), and the optimization stage (b). The framework uses the idea of on-the-fly guidance to integrate the optimizer into the training process. The optimizer will iteratively refine the deformation field predicted by the registration model (for $n$ steps), and the derived optimized deformation field will then be used as pseudo ground truth to train the registration model.}
    \label{fig:overall architecture}
\end{figure*}

\section{Method}\label{sec:method}

\subsection{Overall Structure}\label{sec:optron architecture}

In this work, we present a two-stage training framework with the proposed on-the-fly guidance (OFG), using pseudo-ground truth, it embeds optimization and supervised learning in the training of registration models. \ref{fig:overall architecture} presents the overall structure of the proposed two-stage training framework. 

\textbf{Prediction Stage.} This stage consists of a learning-based registration model. The registration model takes a fixed image $I_{f}$ and a moving image $I_{m}$ and predicts a dense deformation field $\phi$ for each image pair $I_{f}$ and $I_{m}$, i.e.,
\begin{equation}
    F_{\theta}(I_{f}, I_{m}) = \phi
\end{equation}
where $\theta$ denotes the parameters of the registration network. 
Since OFG is a training framework, the prediction stage can utilize any existing learning-based registration model that predicts a deformation field. In our experiments, we used several popular models, such as VoxelMorph \cite{Balakrishnan_2019} in the prediction stage.

\textbf{Optimization Stage.} The optimization stage utilizes the proposed optimizer to iteratively refine the deformation field $\phi_{pre}$ predicted from the current training step by $n$ steps (10 in our default setting). Subsequently, the optimized deformation field $\phi_{opt}$ is used as the pseudo label to provide supervision for the current predicted deformation during the training, forming a feedback loop between the prediction model and the optimizer module.

\subsection{On-the-Fly Guidance Training}\label{method:Tra}
Rather than relying on a fixed pseudo label derived from either a pre-trained model's prediction or the final optimized deformation, our approach introduces on-the-fly guidance. This dynamic supervision evolves alongside the training process to control discrepancies between pseudo labels and current predictions. This approach offers two advantages: \textbf{1)} the limited number (typically 5 to 10) of optimizations incurs acceptable training overhead, \textbf{2)} the optimized deformation serves as an attainable goal for the ongoing training step, providing more direct guidance for the model. In essence, OFG delivers incremental supervision, offering step-by-step guidance for the model. 

The underlying assumption behind OFG is that the model and the optimizer can form a self-improving relationship. The model can provide a reasonable prediction, and in turn, the optimizer can refine that prediction. This is validated in our experiments, as the optimizer can provide a good pseudo label even with random input parameters.

\subsection{Differentiable Deformation Optimizer}\label{method:Opt}
Our on-the-fly guidance hinges on an effective optimizer, we explored three optimization strategies: network-based, downsampled, and our proposed optimizer (see \ref{fig:overall architecture} (b)) with instance optimization and high flexibility for parameter updates, the latter showing the best results (see \ref{tab:ablation}). Detailed comparison in Sec. \ref{sec:ablation_study}. 

The proposed differentiable optimizer is simple yet powerful, taking in the deformation field generated by the prediction model as its initial parameters and optimizing it to generate the pseudo label. It features a Spatial Transformer Network (STN) \cite{STN} without extra parameters, focusing updates solely on the deformation field. During an optimization iteration, the current deformation field is applied to the moving image with the STN, yielding a warped image. An energy function will evaluate the discrepancy between the warped and the fixed image, and the distortion of the deformation field. This loss is backpropagated using Adam \cite{kingma2017adam} or SGD, this essentially refines the deformation field, i.e.,
\begin{equation}
    \phi_{opt}^{(n+1)} = \phi_{opt}^{(n)} - \eta \nabla E_{opt}
\end{equation}
where $n$ denotes the iteration step, $\eta$ is the learning rate, and $\nabla E_{opt}$ represents the gradient of the optimization energy function, implementation detailed in \ref{energy function}.
Notably, the optimizer's role is limited to training, not inference, maintaining training efficiency.

\subsection{Implementation Detail}

\textbf{Training Loss Function.} The model's training utilizes a loss function that enforces supervision from pseudo labels. The model learns by minimizing the discrepancy between the predicted deformation field $\phi_{pre}$, and the optimized deformation field $\phi_{opt}$, which is quantified using MSE. Implemented as follows:
\begin{equation}
    L_{ofg} = \frac{1}{n} \sum (\phi_{pre} - \phi_{opt})^2
    \label{loss-all}
\end{equation}
where $L_{ofg}$ is the model's training loss, it is essentially a MSE-based supervision. Optionally, a weight decay of 0.02 can be added to reduce overfitting for small datasets, as suggested in VoxelMorph \cite{Balakrishnan_2019}. 

\textbf{Optimizer Energy Function.} 
The energy function to be minimized in the differentiable optimizer consists of two terms: an image similarity loss term that captures the difference between the warped image $I_{m }\circ \phi$ and fixed image $I_{f}$ and a $L_2$  regularization loss that imposes smoothness in $\phi$:
\begin{equation}
    E_{opt}(I_{m},I_{f},\phi)=NCC(I_{f}, I_{m}\circ\phi) + \sum_{p\in\Omega}||\nabla\phi(p)||^2
    \label{energy function}
\end{equation}
where $\circ$ is the transform operation which warps $I_{m}$ using $\phi$. The similarity metric we used is the normalized cross-correlation (NCC), $\hat{I}_f(p)$ and $\hat{I}_m(p)$ represent the mean voxel value within a local window of size $n^3$ centered at voxel $p$:
\begin{equation}
    \scriptsize
    NCC(I_f,I_m\circ\phi)=\\
    \sum_{p\in\Omega}\frac{(\sum_{p_i}(f(p_i)-\hat{f}(p))([I_m\circ\phi](p_i)-[\hat{I}_m\circ\phi](p)))^2}{(\sum_{p_i}(f(p_i)-\hat{f}(p))^2)(\sum_{p_i}([I_m\circ\phi](p_i)-[\hat{f}_m\circ\phi](p))^2)}
\label{NCC-function}
\end{equation}

\section{Experiments}\label{sec:exp}

\subsection{Experiment Conditions}

\textbf{Dataset and Preprocessing.} We utilize three public Brain MRI datasets in our study: IXI \cite{ixi}, OASIS \cite{10.1162/jocn.2007.19.9.1498}, and LPBA40 \cite{lpba}, with standard preprocessing steps including skull stripping, resampling, and affine transformation. For IXI, we use 200 volumes for training and 20 for validation; for OASIS, 200 for training and 19 for validation; and for LPBA40, 30 for training, 9 for validation. We also utilize the Abdomen CT-CT dataset \cite{9925717} to evaluate the generalizability of our method on CT registration. 30 for training, and 20 for validation.

\textbf{Evaluation Metrics.} Our evaluation uses two primary metrics: the Dice score (DSC)  \cite{Balakrishnan_2019,BAJCSY19891} for assessing volume overlap in anatomical segmentations, indicating registration accuracy, and the Jacobian matrix to measure deformation field smoothness. The latter involves counting non-background voxels where $\% |J_{\phi}| < 0$, highlighting non-diffeomorphic deformation areas \cite{afas2007neu}.

\textbf{Baseline Methods.} We validated our method based on various popular registration methods. This comparison included two traditional methods, SyN \cite{AVANTS200826} and NiftyReg  \cite{niftyreg} and multiple learning-based methods, VoxelMorph \cite{Balakrishnan_2019},  ViT-V-Net \cite{chen2021vitvnet},  TransMorph \cite{Chen_2022}. All methods are in their default configuration.

\textbf{Experiment Settings.} All models were trained on RTX 4090 for 500 epochs using Adam \cite{kingma2017adam}, with an initial learning rate of 1e-4, batch size of 1, weight decay of 0.02. For the differentiable optimizer, we used an initial learning rate of $0.1$, coupled with a default optimization step count of $10$ during training.

\subsection{Image Registration Results}

\begin{table}[t]
\caption{\small Evaluation results for different methods on various datasets. The OFG architecture provides significant and substantial improvement on the unsupervised learning-based methods. These results validate OFG's effectiveness and generalizability.} 
\label{tab:results}
    \begin{center}
    {\scriptsize{
    \begin{tabular}{llccccc}
    \hline
    Datasets & Methods & Base. DSC $\uparrow$ & \textbf{OFG DSC} $\uparrow$ & Base. $\% |J_{\phi}| < 0$ $\downarrow$ & \textbf{OFG $\mathbf{\% |J_{\phi}| < 0}$} $\downarrow$ \\
    \hline
    \multirow{5}{*}{IXI \cite{ixi}} 
    & SyN \cite{AVANTS200826} & 0.647 & N/A & 1.96e-6 & N/A \\
    & NiftyReg \cite{niftyreg} & 0.585 & N/A & 0.029 & N/A \\
    \noalign{\smallskip}
    \cline{2-6}
    \noalign{\smallskip}
    & VoxelMorph \cite{Balakrishnan_2019} & 0.714 & \textbf{0.737}(+2.3\%) & 1.398 & \textbf{0.516}(-63.1\%) \\
    & ViT-V-Net \cite{chen2021vitvnet} & 0.716 & \textbf{0.738}(+2.2\%) & 1.543 & \textbf{0.545} (-64.7\%) \\
    & TransMorph \cite{Chen_2022} & 0.744 & \textbf{0.760}(+1.6\%) & 1.433 & \textbf{0.794} (-44.6\%) \\
    \hline
    \multirow{5}{*}{OASIS \cite{10.1162/jocn.2007.19.9.1498}} 
    & SyN \cite{AVANTS200826} & 0.769 & N/A & 1.58e-4 & N/A \\
    & NiftyReg \cite{niftyreg} & 0.762 & N/A & 0.011 & N/A \\
    \noalign{\smallskip}
    \cline{2-6}
    \noalign{\smallskip}
    & VoxelMorph \cite{Balakrishnan_2019} & 0.788 & \textbf{0.794}(+0.6\%) & 0.911 & \textbf{0.490} (-46.2\%) \\
    & ViT-V-Net \cite{chen2021vitvnet} & 0.794 & \textbf{0.809}(+1.5\%) & 0.887 & \textbf{0.487} (-45.1\%) \\
    & TransMorph \cite{Chen_2022} & 0.818 & \textbf{0.818}(=) & 0.765 & \textbf{0.517} (-32.4\%) \\
    \hline
    \multirow{5}{*}{LPBA40 \cite{lpba}} 
    & SyN \cite{AVANTS200826} & 0.703 & N/A & 1.18e-4 & N/A \\
    & NiftyReg \cite{niftyreg} & 0.691 & N/A & 1.13e-3 & N/A \\
    \noalign{\smallskip}
    \cline{2-6}
    \noalign{\smallskip}
    & VoxelMorph \cite{Balakrishnan_2019} & 0.658 & \textbf{0.666}(+0.8\%) & 0.288 & \textbf{0.023} (-92.0\%) \\
    & ViT-V-Net \cite{chen2021vitvnet} & 0.663 & \textbf{0.672}(+0.9\%) & 0.390 & \textbf{0.112} (-71.3\%) \\
    & TransMorph \cite{Chen_2022} & 0.678 & \textbf{0.684}(+0.6\%) & 0.438 & \textbf{0.150} (-65.8\%) \\
    \hline
    \end{tabular}
    }}
    \end{center}
\end{table}

\begin{figure}[t]
    \label{fig:results}
    \begin{center}
       \includegraphics[width=0.9\linewidth]{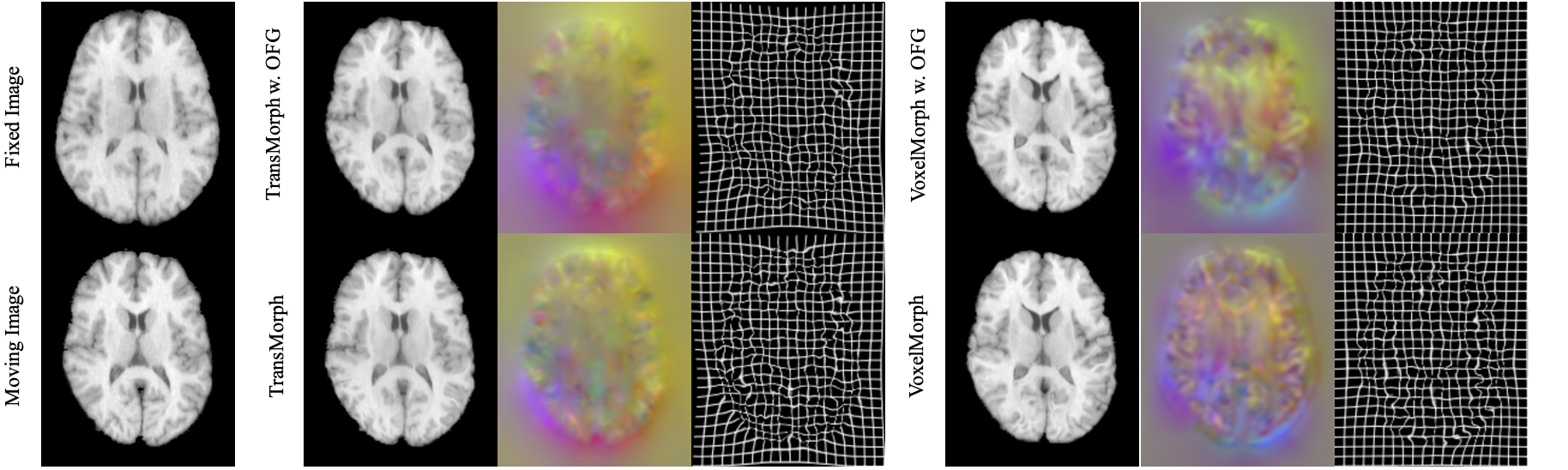}
    \end{center}
    \caption{\small Visualization of registration results on LPBA40 \cite{lpba}. Demo randomly extracted from the comparison results between baseline TransMorph, VoxelMorph (row 2) and their respective model trained with OFG (row 1). OFG shows improved smoothness.}
    \label{fig:vis-results}
\end{figure}

We conducted extensive experiments on the three datasets with three baseline models to showcase the effectiveness and robustness of our proposed framework. Also comparing with the baseline of non-learning methods, see \ref{tab:results}.

\textbf{OFG on Brain MRI.} We have evaluated OFG on brain MRI registration extensively using 3 popular baseline models and datasets. Our method provides a consistent and significant margin on DSC across different models and datasets, demonstrating its effectiveness and generalizability. On the IXI dataset, OFG improved DSC by +1.6\% over TransMorph. For VoxelMorph and ViT-V-Net, it increased DSC by +2.3\% and +2.2\%, respectively, highlighting its general applicability. While on the smaller LPBA40 dataset, OFG's added supervision proved essential in preventing overfitting, underlining the importance of challenging supervision in sparse-data scenarios. Conversely, on the OASIS dataset, OFG showed little improvement on TransMorph, likely due to the dataset's lesser challenge to the model, a hypothesis supported by the lowest training loss of TransMorph on OASIS over all tested cases, suggesting a reduced learning potential for this case. Importantly, for all test cases, our method significantly reduced the percentage of non-diffeomorphic voxels, preventing overly sharp deformations and improving the quality of the registration.

\textbf{OFG on Abdomen CT.} We briefly tested OFG on the Abdomen CT-CT dataset (see \ref{tab:abdo}), with varying optimization configurations. For VoxelMorph, we used MSE as the optimizer energy function, with 3 optimization steps, resulting in a small improvement over baseline. For TransMorph, we used the default energy function, with 5 optimization steps, yielding a much greater improvement over baseline. OFG's effectiveness on another modality showcases its robustness.

\begin{figure}[t]
    \begin{center}
       \includegraphics[width=0.9\linewidth]{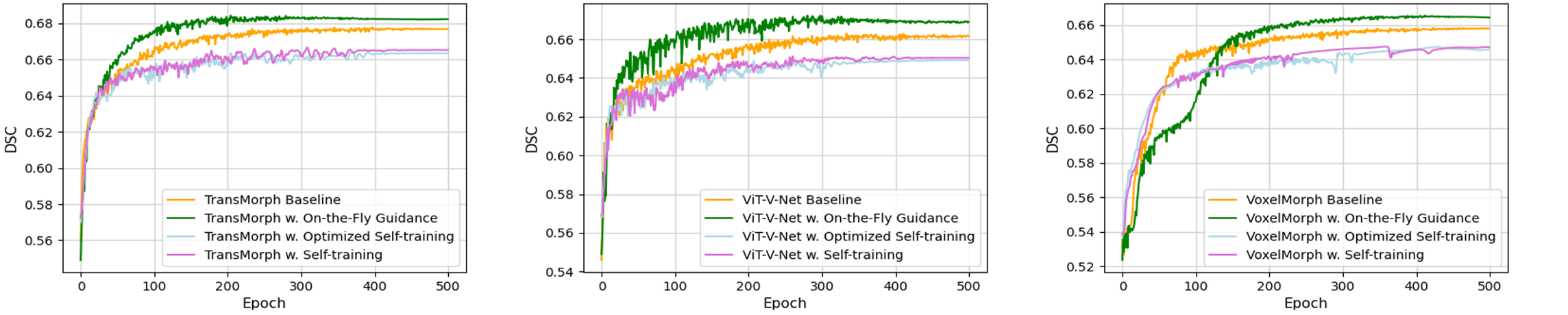}
    \end{center}
    \caption{Visualization comparing training progress and validation DSC on LPBA40 across models. Self-training uses pre-trained network deformation fields as pseudo labels; optimized self-training enhances this with extra optimization steps. Our method achieves the best outcome, with self-training lagging due to convergence complexities.}
    \label{fig:training}
\end{figure}

\textbf{OFG vs. Self-training.} We compared our result with various forms of self-training including Cyclical Self-training (CST). Our method consistently outperforms self-training methods on LPBA40,  see \ref{fig:training}. We also applied CST and OFG on VoxelMorph, and tested on LPBA40, our method provides +3.8\% better DSC while halving the Jacobian, see \ref{tab:cst_vs_ofg}. This is largely due to the OFG training strategy explained in \ref{method:Tra}.

\begin{table}[t]
\begin{minipage}[b]{0.59\linewidth}
\centering
\begin{scriptsize}
\caption{Abdomen CT registration. VoxelMorph uses 3-step MSE optimizer, TransMorph uses 5-step NCC optimizer.}
\label{tab:abdo}
\begin{tabular}{lcccc}
    \hline
    Methods & DSC & Jacob. & \textbf{OFG DSC} & \textbf{OFG Jacob.} \\
    \hline
    VoxelMorph & 0.312 & 0.165 & \textbf{0.319}(+0.7\%) & \textbf{0.120}(-27.3\%)\\
    TransMorph & 0.317 & 0.165 & \textbf{0.382}(+6.5\%) & \textbf{0.010}(-93.9\%)\\
    \hline
\end{tabular}
\end{scriptsize}
\end{minipage}
\begin{minipage}[b]{0.39\linewidth}
\centering
\begin{scriptsize}
\caption{Cyclical Self-training vs. OFG on LPBA40. OFG performs significantly better.}
\label{tab:cst_vs_ofg}
\begin{tabular}{lcc}
    \hline
    Methods &  DSC & Jacob. \\
    \hline
    VoxelMorph w. CST & 0.628 & 0.047 \\
    VoxelMorph w. OFG & \textbf{0.666} & 0.023 \\
    \hline
\end{tabular}
\end{scriptsize}

\end{minipage}
\end{table}

\subsection{Ablation Study}
\label{sec:ablation_study}

\begin{figure}[t]
\scriptsize
    \begin{center}
       \includegraphics[width=0.9\linewidth]{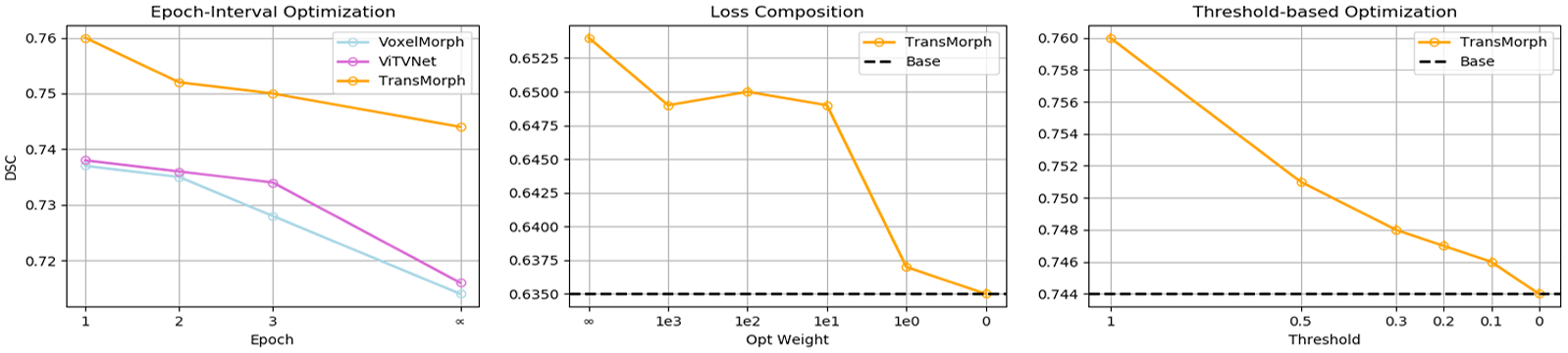}
    \end{center}
    \caption{Ablation results on blending OFG with baseline unsupervised learning. From left to right, we evaluated optimization frequency, loss blending, and probabilistic optimization, generally showing a decrease in performance when the intensity of OFG is decreased, proving its effectiveness.}
    \label{fig:mix}
\end{figure}

\begin{table}[t]
\begin{center}
\begin{minipage}[b]{0.51\linewidth}
\centering
{\scriptsize{
    \caption{Ablation results of the OFG Optimizer on LPBA40. This evaluation focuses on the initial 30 epochs.}
    \label{tab:dopt}
    \begin{tabular}{lccc}
    \hline
    Designs &  DSC & Opt. Time & VRAM \\
    \hline
    None & 0.635 & N/A & 12.07 GB \\
    Network-based & 0.642 & 13.604 & 14.35 GB \\
    DownSample & 0.610 & \textbf{0.097} & 12.44 GB \\
    \hline
    \textit{Ours} & \textbf{0.654} & 3.228 & 13.07 GB \\
    \hline
    \end{tabular}

}}
\end{minipage}
\begin{minipage}[b]{0.48\linewidth}
\centering
{\scriptsize{
    \caption{Comparison between our optimizer and network-based optimizer on IXI. Results for first 200 epochs.}
    \label{tab:ablation}
    \begin{tabular}{lccc}
    \hline
    Designs & DSC & $\% |J_{\phi}| < 0$ & Opt. Time \\
    \hline
    VoxelMorph-1 & 0.691 & 0.456 & 1.174 \\
    VoxelMorph-2 & 0.690 & 0.383 & 2.314 \\
    VoxelMorph-5 & 0.693 & 0.385 & 5.856 \\
    \hline
    \textit{Ours} & \textbf{0.731} & 0.503 & 3.018 \\
    \hline
    \end{tabular}
}}
\end{minipage}
\end{center}
\end{table}

\textbf{OFG Intensity Ablation.} To show OFG's deciding factors and how they influence performance, we blended OFG with baseline unsupervised loss in 3 different forms: \textbf{1)} Optimization frequency: only use optimizer every $n$ epochs, i.e., decreased optimization frequency. \textbf{2)} Loss weight composition: adding NCC loss into the loss function, i.e., $L = \alpha L_{ofg} + \beta L_{NCC}$. \textbf{3)} Probabilistic optimization: only randomly optimizes a portion of the image instances during training. As shown in Fig \ref{fig:mix}, we observed a decrease in performance when the intensity of OFG decreases, in all three forms. Notably, a low optimization frequency resembles the training strategy used in Cyclical Self-training. This result suggests a higher optimization frequency (intensity) provides improved performance.

\textbf{Optimizer Design.} We also evaluated 2 other optimizer designs for our framework, including: \textbf{1)} Network-based optimizer: using a network capable of fitting a general transformation to optimize deformation fields, in our case, we used $n$ cascaded VoxelMorph. \textbf{2)} Downsampled optimizer: to improve the computational overhead, a downsampled optimizer reduces all dimensions in half, with only 1/8 of the updatable parameters. Table \ref{tab:dopt} and \ref{tab:ablation} show the proposed design achieves the best performance.

\textbf{Optimization Steps.} We assessed the impact of optimization steps ranging from 1 to 15 on training outcomes to balance computational efficiency with optimization quality. Findings indicate that 5 to 10 steps offer optimal balance, enhancing optimization quality without significantly lengthening training time (only with a 10 to 18\% increase), with no notable benefits from exceeding this range. See Fig. 1. in supplementary material for detailed results.

\textbf{Self-improving Relationship.} OFG is based on the concept that the model and optimizer can enhance each other, with the optimizer's robustness being key to generating high-quality pseudo labels in various scenarios. Our findings indicate that the optimizer can effectively refine initial deformations, even those generated randomly or from models with random initialization, leading to significant improvements. See Fig. 2. in supplementary material for detailed results.

\section{Conclusion}

This work introduces On-the-Fly Guidance, a training framework that successfully applies supervised-style training to learning-based registration models. Demonstrating significant improvements on benchmark datasets, especially with deformation smoothness, OFG has proven its effectiveness and generalizability. OFG only comes with limited training overhead and no inference overhead. The flexibility of our method allows future work to focus on aspects such as improving the efficiency of the optimizer, using dynamic optimization steps, altering the optimizer design, and so on.

\begin{credits}

\subsubsection{\discintname}
The authors have no competing interests to declare that are relevant to the content of this article.
\end{credits}

\newpage
\bibliographystyle{splncs04}
\bibliography{Paper-0519}



%
\title{Supplementary Material}

\author{
Yuelin Xin\inst{1,2}*
\and
Yicheng Chen\inst{3,5}*
\and
Shengxiang Ji\inst{4,6}*
\and
\\Kun Han\inst{2}*
\and
Xiaohui Xie\inst{2}
}

\authorrunning{Y. Xin et al.}

\institute{
\mbox{
\inst{1}University of Leeds
\quad
\inst{2}University of California, Irvine
\quad
}
\mbox{
\inst{3}Tongji University
\quad
\inst{4}Huazhong University of Science and Technology
}
\mbox{
\inst{5}Fudan Univeristy
\quad
\inst{6}University of California, San Diego
}
}

\maketitle

\begin{figure}
    \begin{center}
       \includegraphics[width=0.9\linewidth]{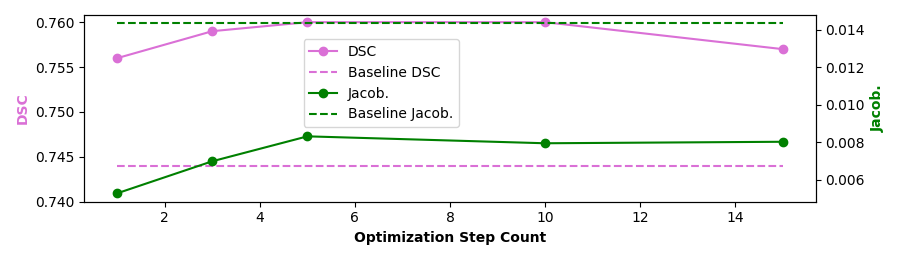}
    \end{center}
    \vskip -0.4cm
    \caption{Ablation on optimization steps (TransMorph on IXI). Results show that 5 to 10 steps offer optimal balance, with no notable benefits from exceeding this range. Thus, we recommend to use 5 to 10 steps.}
    \label{fig:step}
\end{figure}

\begin{figure}
    \begin{center}
       \includegraphics[width=0.9\linewidth]{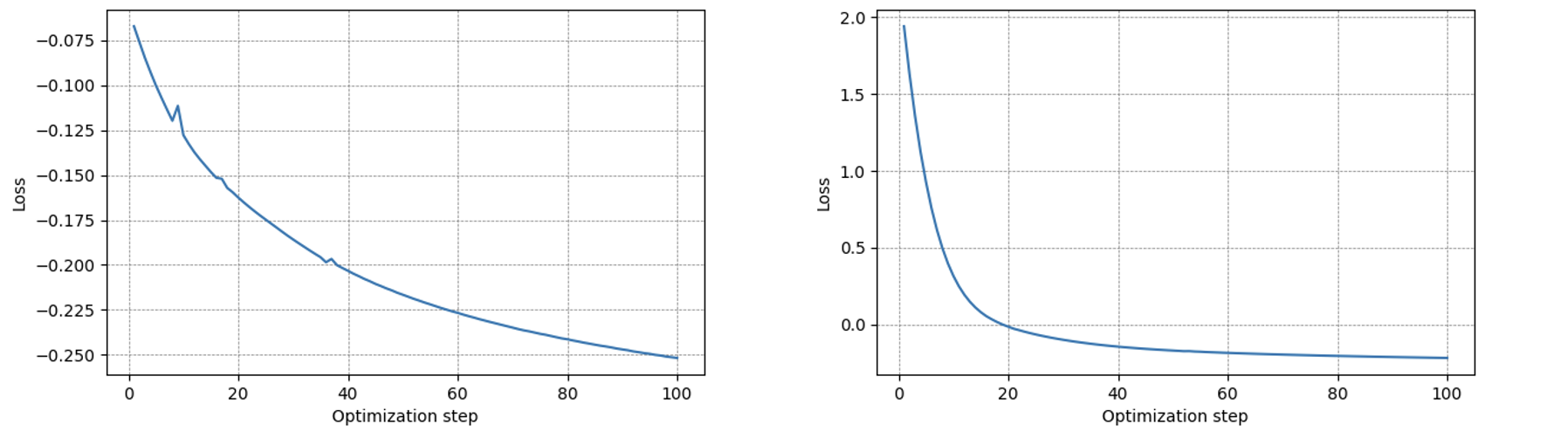}
    \end{center}
    \vskip -0.4cm
    \caption{The optimizer can quickly and effectively refine the deformation field even deformation from models with random initialization (left) or random parameters (right). With DSC increasing from 0.4260 to 0.5436 (left), and 0.4260 to 0.5101 (right).}
    \label{fig:step}
\end{figure}

\begin{figure}[h]
    \begin{center}
       \includegraphics[width=0.9\linewidth]{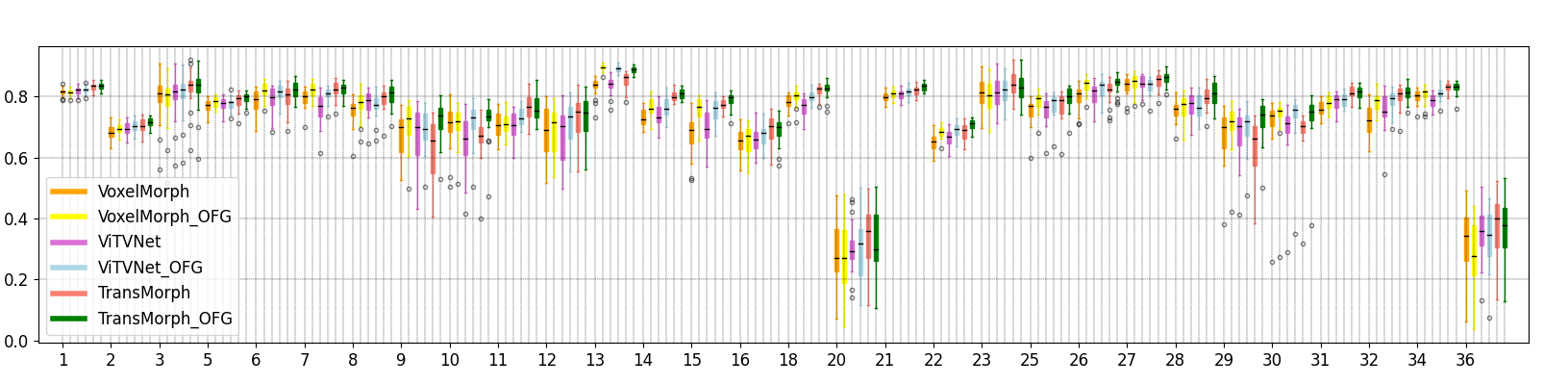}
    \end{center}
    \vskip -0.4cm
    \caption{Evaluation results for each label and different methods on IXI. OFG provides significant and substantial improvement on the unsupervised learning-based methods for most labels, it also surpasses the self-training and optimized self-training method.}
    \label{fig:results_per_label}
\end{figure}

\begin{figure}
    \begin{center}
       \includegraphics[width=0.8\linewidth]{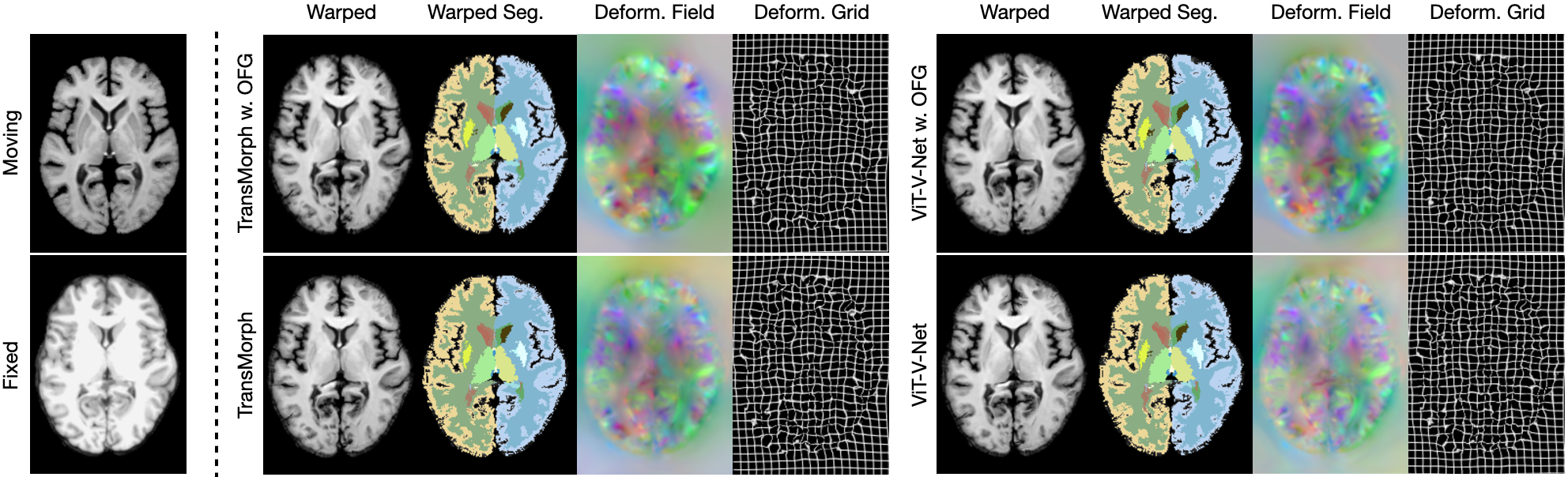}
    \end{center}
    \vskip -0.5cm
    \caption{Registration results comparisons on IXI. Demo randomly extracted from the comparison results between baseline TransMorph, ViT-V-Net (row 2) and their respective model trained with OFG (row 1). OFG shows improved smoothness.}
    \label{fig:results_additional}
\end{figure}

\begin{figure}
    \begin{center}
       \includegraphics[width=0.8\linewidth]{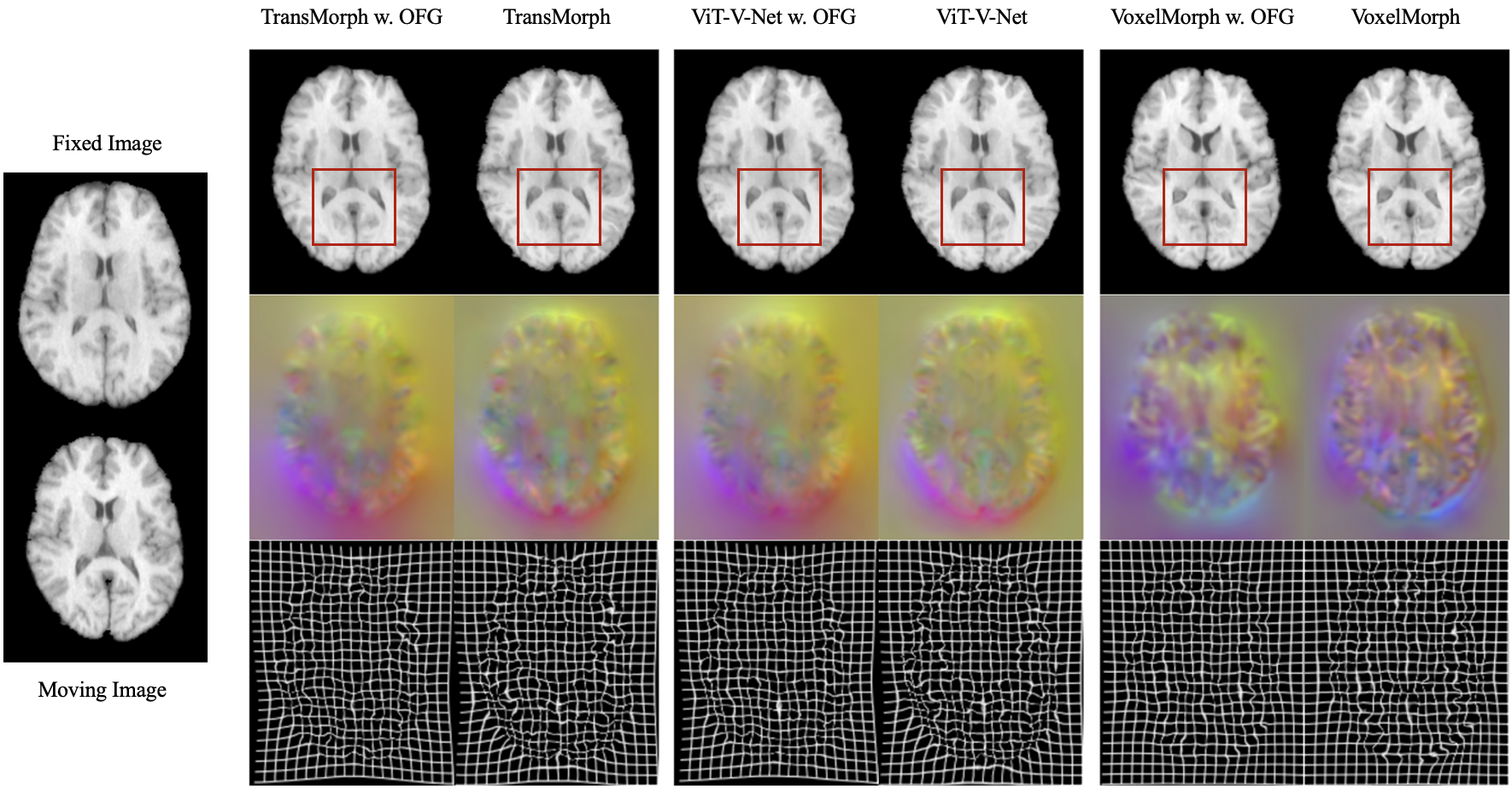}
    \end{center}
    \vskip -0.5cm
    \caption{Registration results comparisons on LPBA40. The red bounding box outlines a region in which we can easily compare the difference between registration outcomes. We can also observe that the deformation fields are smoother for models with OFG.}
    \label{fig:results_additional}
\end{figure}

\begin{figure}
    \begin{center}
       \includegraphics[width=0.8\linewidth]{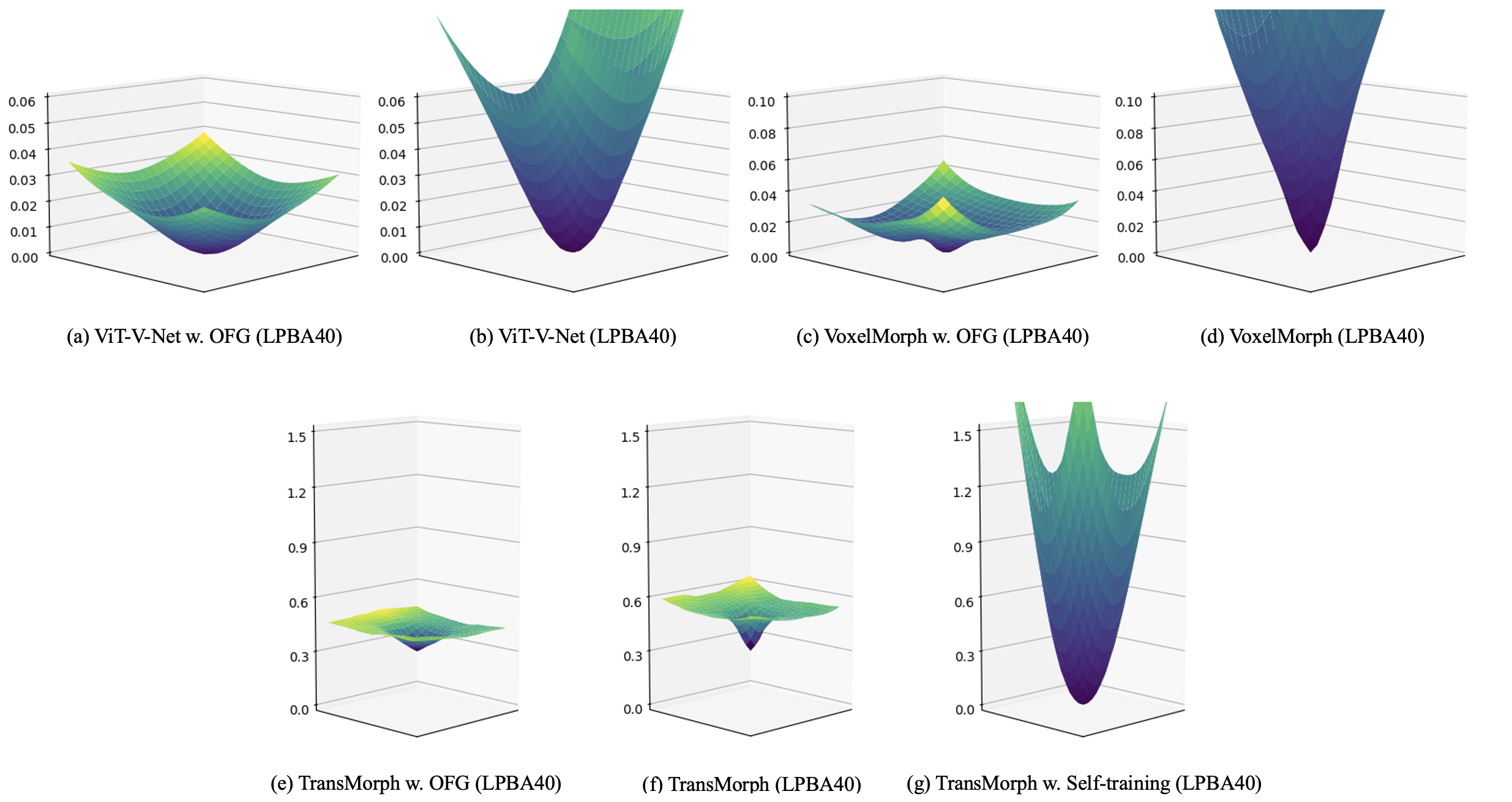}
    \end{center}
    \vskip -0.5cm
    \caption{Loss landscape visualization for comparing model trained with and without OFG, showing OFG with significant improvement in loss landscapes for ViT-V-Net, VoxelMorph and TransMorph on LPBA40. In Addition, for the comparison of TransMorph, we also illustrated the landscape for self-training, which is significantly less smooth compared with OFG.}
    \label{fig:landscape_lpba}
\end{figure}


\end{document}